\documentclass[conference]{IEEEtran}
\IEEEoverridecommandlockouts
\usepackage{cite}
\usepackage{url}
\usepackage{todonotes}
\usepackage{amsmath,amssymb,amsfonts}
\usepackage{algorithmic}
\usepackage{graphicx}
\usepackage{textcomp}
\usepackage{xcolor}
\def\BibTeX{{\rm B\kern-.05em{\sc i\kern-.025em b}\kern-.08em
    T\kern-.1667em\lower.7ex\hbox{E}\kern-.125emX}}
\begin{document}

\title{A novel Network Science Algorithm for Improving Triage of Patients.
\thanks{Identify applicable funding agency here. If none, delete this.}
}

\author{\IEEEauthorblockN{Pietro Hiram Guzzi}
\IEEEauthorblockA{\textit{Dept. Medical and Surgical Sciences} \\
\textit{University of Catanzaro}\\
Catanzaro, Italy \\
 ORCID: 0000-0001-5542-2997}
\and
\IEEEauthorblockN{Annamaria Defilippo}
\IEEEauthorblockA{\textit{Dept. Medical and Surgical Sciences} \\
\textit{University of Catanzaro}\\
Catanzaro, Italy \\
email address or ORCID}
\and
\IEEEauthorblockN{Pierangelo Veltri}
\IEEEauthorblockA{\textit{DIMES} \\
\textit{UNICAL}\\
Rende, Cosenza\\
pierangelo.veltri@unical.it}
}

\maketitle

\begin{abstract}
Patient triage plays a crucial role in healthcare, ensuring timely and appropriate care based on the urgency of patient conditions. Traditional triage methods heavily rely on human judgment, which can be subjective and prone to errors. Recently, a growing interest has been in leveraging artificial intelligence (AI) to develop algorithms for triaging patients. This paper presents the development of a novel algorithm for triaging patients. It is based on the analysis  of patient data to produce  decisions regarding their prioritization. The algorithm was trained on a comprehensive data set containing relevant patient information, such as vital signs, symptoms, and medical history. The algorithm was designed to accurately classify patients into triage categories through rigorous preprocessing and feature engineering. Experimental results demonstrate that our algorithm achieved high accuracy and performance, outperforming traditional triage methods. By incorporating computer science into the triage process, healthcare professionals can benefit from improved efficiency, accuracy, and consistency, prioritizing patients effectively and optimizing resource allocation. Although further research is needed to address challenges such as biases in training data and model interpretability, the development of AI-based algorithms for triaging patients shows great promise in enhancing healthcare delivery and patient outcomes.
\end{abstract}


\section{Introduction}
\label{sec:introduction}

Management of the access of people to emergency departments (ED) is a critical problem for healthcare administration. The correct management of queues may improve the level of quality of hospitals as well as  may contribute to control costs and reimbursement. 

Access to ED presents some peculiar characteristics in different countries. For instance, in Italy, the request for access to ED is universal, so all the people staying in Italy, without any restriction, may require access. This often cause overcrowding, even in absence of special conditions such as pandemics, so the problem is high important \cite{parmeggiani2010healthcare,savioli2022emergency,bambi2021new}. 

The management of queues is based on prioritising patients considering the clinical aspects to guarantee high-priority access to patients requiring immediate (or urgent) care. For these reasons, a set of systems, known as triage systems have been introduced. Triage systems implement patient screening to prioritise examination at an adequate priority level. 

In Italy, triage is a process by nursing staff before entering the treatment rooms, and it is ruled by the National Ministry of Health \url{https://www.salute.gov.it/imgs/C_17_notizie_3849_listaFile_itemName_1_file.pdf}. Still, the system's core is a four-level in-hospital triage based on an acuity scale measurement and it may present some little difference among regions.

Considering a worldwide scenario, strategies adopted in EDs are based on the  \textit{streaming}, which is based on grouping patients based on the severity of illness and the assignment of patients into separate areas of the ED (e.g. fast-track, see and treat minor injuries) \cite{meyer2013physicians}.

ED triage systems are diffused worldwide, and among the others, according to literature \cite{bambi2021new}, the most diffuse and adopted systems are the Australasian Triage System (ATS) \cite{putri2020australasian}, the Emergency Severity Index (ESI, United States) \cite{wuerz2001implementation}, the Manchester Triage System (MTS) \cite{azeredo2015efficacy}, and the Canadian Triage and Acuity Scale (CTAS) \cite{j2003canadian,bullard2017revisions}. In parallel, some other indexes are currently in use, such as the Korean Triage and Acuity Scale  (KTAS) \cite{kwon2019korean}, the Taiwan Triage Acuity Scale  (TTAS) \cite{ng2011validation},  and the South African Acuity Scale (SAAS) \cite{meyer2018validity}.

Machine Learning and Artificial intelligence (AI) algorithms may create predictive clinical models able to handle large heterogeneous datasets, such as electronic medical records (EMRs) \cite{cheung2019machine,canino2015analysis}. In particular, AI models provide better prediction tasks, outperforming classical clinical scoring systems \cite{hinson2019triage}. Consequently, several prediction models have been developed to improve the triage process providing stratification of patients as well as more sensitivity and specificity to the proper identification of patients at greater risk of mortality. 
In this study, we present a novel clinical algorithm based on artificial intelligence and network science for assigning priority to patients.

\begin{figure}
    \centering
\includegraphics[width=0.5\textwidth,height=0.6\textheight]{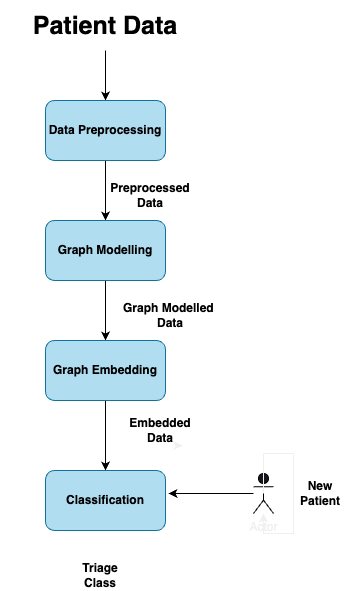}
    \caption{Caption}
    \label{fig:patientflow}
\end{figure}

Figure \ref{fig:patientflow} depicts these steps. We start by considering clinical patient data extracted from patient records. These data include analytical observation as well as subjective ones. After an initial preprocessing step to identify possible noise and outlier a modelling phase is applied. In this step each patient is modelled as a node of a graph, while edges represent the similarity among the observation data. At this point graph is embedded into a latent space. Finally patient are classified into the risk groups by applying a node classification algorithm. Since the embedding is inductive, when a novel patient is available, a novel node is added to the graph and embedded.

We tested our pipeline on public data to demonstrate the effectiveness of our approach and the improvement with respect to state of th e art approaches.

\section{Related Work}
\label{sec:related}

Triage systems for emergency departments  are diffused worldwide \cite{defilippo2023computational}. Some of the most diffuse and adopted systems are the Australasian Triage System (ATS) \cite{putri2020australasian}, the Emergency Severity Index (ESI, United States) \cite{wuerz2001implementation}, the Manchester Triage System (MTS) \cite{azeredo2015efficacy}, and the Canadian Triage and Acuity Scale (CTAS) \cite{j2003canadian,bullard2017revisions}.

ESI is one of the four most widely used classification systems internationally and is often used to learn and test machine learning-based risk prediction systems. It uses five levels to stratify patients on the basis of the vital conditions and the potential threat to the patient's life or vital and non-vital organs. The operator assigns the appropriate level by choosing between maximum urgency (ESI level 1 or 2), minimum speed (ESI level 4 or 5) and intermediate urgency level (ESI 3) \cite{wuerz2001implementation}.

Despite the worldwide adoption and diffusion, some common problems affect the triage systems:  such as dependence on subjective medical staff assessment and the possibility of having many missing variables. Consequently, the possibility to support human decision has been explored in the past, as summarised in  \cite{levin2018machine}.

Automatic, or computational based triage systems, presents some advantages such as: (i) stability of assignment \cite{levin2018machine}, (ii) filtering out noise in the patient variables, (iii) model and analyse patient similarity,  (i.e. by modelling the set of patients in a network which evidences patient similarity \cite{yu2020machine,levin2018machine,choi2019machine,kim2021automatic}; (iv) avoiding  patients under triaged into low severity levels \cite{inokuchi2022machine}; (vi) avoid  of racial, gender, age bias \cite{allen2020racially}. 

\section{The Proposed Method}


The data-set for the analysis was found on the Kaggle platform ( \url{https://www.kaggle.com/datasets/hossamahmedaly/patient-priority-classification} ).


The data-set is composed of 6962 instances and 16 features excluding the target feature. 
Each instance is a patient described by the features that are referred to patient's symptoms in order to identify their conditions and assign the correct triage score. In particular, the following information is taken into consideration :
 \begin{itemize}
\item \textit{Age}: age of the patient.
\item \textit{Gender}: patient's sex.
\item \textit{Chest pain type}: the type of chest pain.
\item \textit{Blood pressure}: blood pressure value.
\item \textit{Cholesterol}: cholesterol level.
\item \textit{Max heart rate}: maximum heart rate value.
\item \textit{Exercise angina}: presence of angina.
\item \textit{Plasma glucose}: glucose level in blood plasma.
\item \textit{Skin thickness}: any thickening of the skin.
\item \textit{Insulin}: insulin level.
\item \textit{BMI}: body mass index obtainable from the square of the ratio between the patient's weight and patient's height.
\item \textit{Diabetes Pedigree}: genetic predisposition to diabetes.
\item \textit{Hypertension}: elevated blood pressure. 
\item \textit{Heart disease}: presence of heart disease.
\item \textit{Residence type}: type of residence place.
\item \textit{Smoking status}: defines whether or not the patient is a smoker.
\end{itemize}

There are two categorical features that will be pre-processed in the next phase and their unique values are the following:
\begin{itemize}
\item \textit{Residence type}: ['Urban' 'Rural']
\item \textit{Smoking status}: ['never smoked' 'smoke' 'previously smoked' 'Unknown']
\end{itemize}

The output of the classifier is a multilevel categorical feature which has four levels corresponding to four triage levels.

In that case, the values associated with the labels have the following meaning:
\begin{itemize}
\item \textit{Red}: needs immediate attention for a critical life-threatening injury or illness; absolute priority and transport first for medical help.
\item \textit{Orange}: risk for  vital functions if no action is taken in a short time. The intensity of the symptom and the vital parameters correlated to the symptom are evaluated to give immediate attention. 
\item \textit{Yellow}: urgent condition that can be deferred without evolutionary risk and without compromising vital functions. It is advisable to do a   re-evaluation when the waiting time becomes excessively long.
\item \textit{Green}: condition with minor urgency because there are no alterations of vital functions and no critical symptoms. 
\end{itemize}

The data-set is unbalanced, but it is common in the literature, because it can be seen that many patients are classified with the intermediate level of urgency. It is justified by the need to prevent the underestimation of  the symptoms reported (under-triage), but at the same time, to avoid  unreal emergencies (over-triage) that can subtract resources. It is explained below, in this experimental phase, how to solve this drawback with oversampling and undersampling techniques.

Initially, we preprocessed the dataset by applying following steps.

\begin{enumerate}
    \item \emph{Management of duplicates and null values}: We deleted all the rows with null values and  As a result, the number of instances of the dataset became 6551.
    \item \emph{Management of inconsistent values}: a single inconsistent value, equal to unknown, was identified in 1,886 instances in correspondence with the characteristic \textit{ smoking status}. In this case, on the basis of the statistical distribution of the considered feature, since it is not a normal distribution, the values were filled with the value of the mode, calculated on the rest of the data.
    \item \emph{Label Encoder for categorical features}: Dataset contains three categorical features: \textit{Residence type}, \textit{Smoking status} and  the third is the target feature.
    \item \emph{Oversampling and Undersampling}: It is necessary to carry out an oversampling phase of the less represented classes for the unbalanced dataset. SMOTE (Synthetic Minority Oversampling Technique)  \cite{chawla2002smote} is one of the most commonly used oversampling methods to deal with class imbalance. It is based on using a k nearest neighbours to create synthetic data. 
    \item \emph{Normalization with Min-Max Scaler}: This last pre-processing phase had the aim of normalizing the feature values in the range $[0,1]$, i.e. with a maximum value equal to 1 and a minimum value equal to 0. The idea is to unify the different scales of the values, in order to contribute fairly to the training of the model, avoiding the creation of bias. Normalization can be done thanks to Scikit-learn's MinMaxScaler module. It should be noted that normalisation was carried out on all features except on the target.
    
\end{enumerate}
\begin{figure*} [h]
    \centering
\includegraphics[width=0.7\textwidth,keepaspectratio]{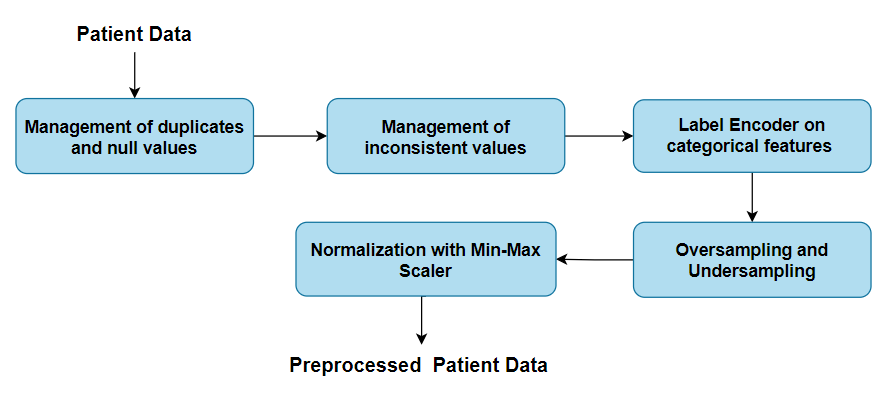}
    \caption{Preprocessing flow}
    \label{fig:flow}
\end{figure*}

\begin{figure*} [h]
    \centering
\includegraphics[width=0.7\textwidth,keepaspectratio]{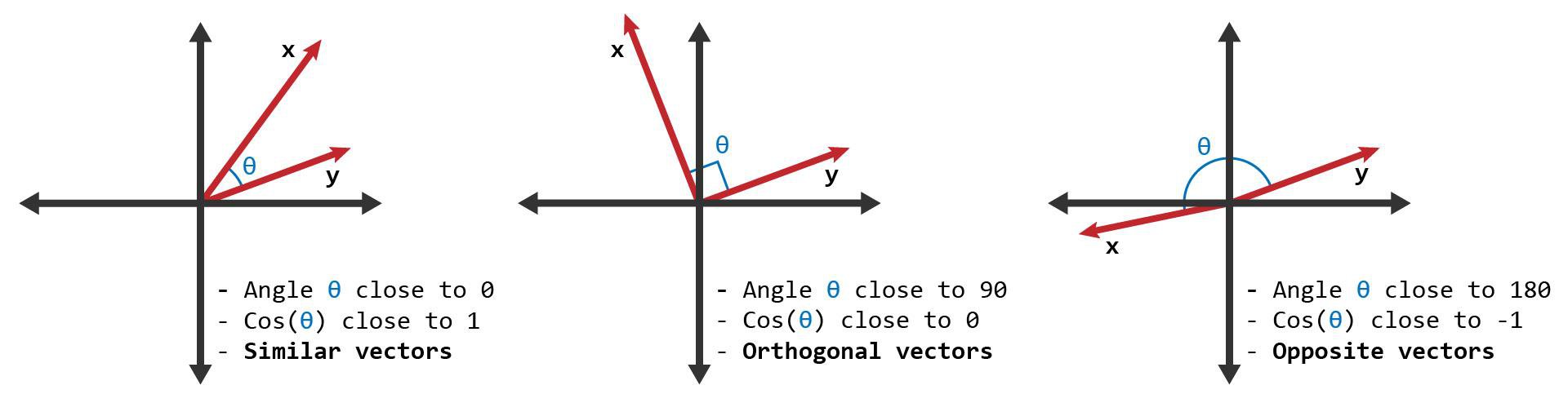}
    \caption{Meaning of cosine similarity}
    \label{fig:cosine_similarity}
\end{figure*}

\subsection{Graph Modelling}

After the initial preprocessing we build the graph representing patient and their similarity. In such a model, the nodes represent the patients, while the weighted edges evaluate their similarity considering their clinical features. Similarity values range from 0 (i.e., no edge) to 1, with a maximum similarity. 


Similarity is calculated using the measure of cosine similarity between instances \cite{teji2023graph,gu2022modeling} and the two metrics relating to the Euclidean distance and the Manhattan distance between instances.

The Scikit-learn module offers the \textit{sklearn.metrics.pairwise} sub-module which provides functions dedicated to calculate some metrics for measuring distance and others for measuring similarity between data-set elements. 
It should be noted that the cosine similarity provides how similar the instances are to each other \cite{guzzi2012semantic}. Similarity values range from 0 (i.e., no edge) to 1 (i.e., maximum similarity).  On the other hand, distance measurements show how dissimilar the instances are, because if the distance value is high, the patients are different from each other \cite{teji2023graph,gu2022modeling}.  

The \textit{nx.Graph} function of NetworkX allows to create three empty graphs which will be populated with the instances of the data-set. Furthermore, each node has the values of the features as attributes. 

For each considered metric, a specific graph creation function was defined as follows: 
\begin{itemize}
\item the link  between the nodes is created only if the value of cosine similarity is higher than the settled threshold. 
\item the link  between the nodes is created only if the value of Euclidean distance is lower than the settled threshold, similarly for the Manhattan distance. 
\end{itemize}

In addition, each link will have a weight that corresponds to the similarity or distance value. In order to set appropriate thresholds, the average values of similarity and distances on the data-set were calculated before, to choose the values that would neither cause the exclusion of too many nodes nor the creation of an excessive number of links. 


\subsection{Graph Embedding}

Machine learning algorithms have been successful in extracting meaningful information from data \cite{teji2023graph,guzzi2022editorial,roy2023graph}. However, they are only effective when the input data is organized in tabular structures. Graph data, on the other hand, does not have a tabular structure. Therefore, graph embedding approaches are the best way to convert a graph structure into a tabular one that can be mined with deep learning algorithms. Graph Representation Learning (GRL) or graph embedding is a set of algorithms and methods to encode graph structures into low-dimensional spaces, using both mathematical and soft-computing techniques \cite{hamilton2017inductive,guzzi2023analysis}. The goal of GRL methods is to learn a mapping of each node into a subspace with a smaller size while preserving as much information as possible. Then, the analysis that is performed in the latent subspace can be translated back into the graph space to help solve relevant problems. A common way to perform GRL is to embed each node into a separate point in a subspace while preserving the initial distance between nodes, i.e. similar nodes should be embedded into close points. Node embedding methods can be divided into three major categories: i) Matrix-Factorization; ii) Random-walks; and iii) Graph Neural Networks. 

The proposed approaches belonging to the first two classes have two main drawbacks: (i) they do not take into account data related to nodes, (i.e. node features); (ii) they are inherently transductive, so they need to recalculate the whole embedding in case of any graph modification (e.g. node/edge insertion or removal). To address these issues, a variety of different approaches have been developed.

Consequently we here adopted three methods belonging to the third class, available in NetworkX package: Graph Convolutional Networks \cite{betkier2023pocketfindergnn}, GATv2Conv \cite{brody2021attentive}, and GraphSage \cite{hamilton2017inductive}.

The first architecture is composed by five GCNConv layers of 64 nodes, and it is used for the analysis of the graphs created on the basis of the cosine similarity (as represented in Figure \ref{fig:cosine_similarity}) and the Manhattan distance. 
On the other hand, for the graph created based on the Euclidean distance, four GCNConv layers of dimension equal to 32 are the network's structure. The input size is equal to 16 (in agreement with the number of features), while the output equal to 4 (in agreement with the number of feature target's classes). 

In Graph Attention Networks, we defined  the size of the input samples (i.e. 16) and the output to be obtained (i.e. 4), in the GATv2Conv layer, the number of heads must be defined as the number of attention heads to implement the mechanism of attention.

In that case, the multi-attention mechanism makes use of 4 heads in the model for the two layers that make up the network. 

The architecture of GraphSAGE is composed by 5 SAGEConv layers each with the input/output dimensions respectively of: 16, 64 ; 64, 32 ; 32, 16 ; 16, 8 ; 8.4. The ‘max’ aggregator is the max pooling operation used for each layer, with the exception of the third where it was applied the mean-based aggregation.

\section{Results}
After that explanation of the methods applied, it follows that the results will be reported in the this section. The purpose is to discuss which are the performances in predicting triage score that could be obtained with each architecture of GNN applied to a specific type of networks. It is useful to remember that each architecture was trained and tested on each graph created by the considered measures. 


The GCN architecture and the GAT architecture were trained for 300 epochs on the graphs created by the Manhattan distance and the Cosine similarity; on the contrary, they were trained for 200 epochs on the graph created by the Euclidean distance. Instead, the GraphSage architecture was trained for 300 epochs for all graph structures. 
In the table below, the results are illustrated and compared, showing both which graph structure was tested and which GNN architecture was used. According to this table, GraphSage could be considered the best architecture to complete the task on all the structures.
\begin{figure*} [h]
    \centering
\includegraphics[width=0.7\textwidth,keepaspectratio]{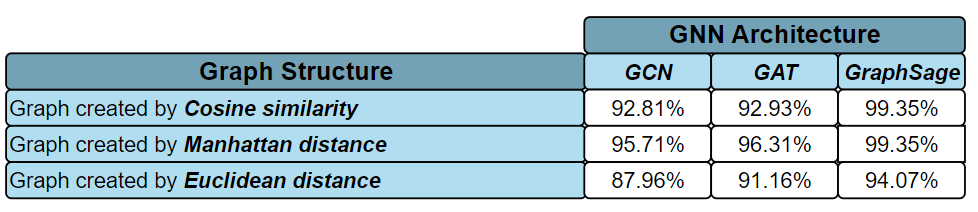}
    \caption{Test accuracy}
    \label{fig:flow}
\end{figure*}

\subsection{Comparison with respect classification on tabular data}
\begin{figure*}
    \centering
\includegraphics[width=0.55\textwidth,keepaspectratio]{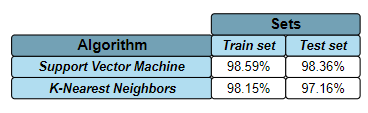}
    \caption{Accuracy on tabular data}
    \label{fig:flow}
\end{figure*}
It could be useful to compare the results obtained with networks related to  classification on tabular data. Therefore, to do that, it was trained two types of algorithms: Support Vector Machine and K-Nearest Neighbors on the same patient data used for classification on graphs. Similarly, the same pre-processing phase was carried out. $\\$
The tabular data were splitted in train, test and evaluation set with a proportion of 30\% for test set. In addition, a percentage of 30\% of the test set was used for evaluation. 
The results obtained (shown in Figure 3) were higher than GCN and GAT applied on graph structures but are lesser than GraphSage results that could be improved in future research to obtain better results.


\section{Conclusion}
The work addresses a critical issue in healthcare administration: the management of access to emergency departments (EDs). The study focuses on Italy, where universal access to EDs often leads to overcrowding, highlighting the importance of effective queue management. The paper introduces various global triage systems, emphasizing the need for accurate prioritization to ensure immediate care for critical patients.

The authors propose a novel approach utilizing machine learning and network science to assign priority to ED patients. They preprocess patient data, create graphs based on similarity metrics (cosine similarity, Euclidean distance, and Manhattan distance), and apply Graph Neural Networks (GNNs) including GCN, GAT, and GraphSage for classification. The methods are rigorously described, demonstrating a thorough understanding of both the medical context and the applied techniques.

Strengths:

Comprehensive Approach: The work offers a comprehensive analysis of the problem, including the exploration of various triage systems, preprocessing techniques, and GNN architectures. This holistic approach ensures a well-rounded study.

Innovative Methodology: The use of GNNs for triage classification, especially GraphSage, showcases innovation. The incorporation of network science principles into healthcare decision-making is a novel and promising approach.

Data Handling: The study demonstrates a sound understanding of data preprocessing, handling missing values, and managing categorical variables. Addressing these issues appropriately is critical for accurate machine learning models.

Comparison with Baseline: Comparing GNN-based classification with traditional machine learning algorithms on tabular data provides valuable insights, enabling a comprehensive evaluation of the proposed approach.

 \section{Acknowledgements}
 This work was funded by the Next Generation EU - Italian NRRP, Mission
4, Component 2, Investment 1.5, call for the creation and strengthening of
'Innovation Ecosystems', building 'Territorial R\&D Leaders' (Directorial
Decree n. 2021/3277) - project Tech4You - Technologies for climate change
adaptation and quality of life improvement, n. ECS0000009. This work
reflects only the authors' views and opinions, neither the Ministry for
University and Research nor the European Commission can be considered
responsible for them.

\end{document}